%% file: acl2020.tex
\documentclass[11pt,a4paper]{article}
\usepackage[hyperref]{acl2020}
\usepackage{times}
\usepackage{latexsym}

\usepackage{microtype}

\aclfinalcopy 
\def\aclpaperid{\#914} 

\usepackage{natbib}
\usepackage{graphicx}
\usepackage{multirow}
\usepackage{amsmath}
\usepackage{booktabs}
\usepackage{subfigure}
\usepackage{epstopdf}
\usepackage{comment}
\usepackage{amssymb}
\usepackage{url}
\usepackage{xcolor}

\title{A Novel Cascade Binary Tagging Framework for \\ Relational Triple Extraction}

\author{Zhepei Wei\textsuperscript{\rm 1,2}, Jianlin Su\textsuperscript{\rm 4}, Yue Wang\textsuperscript{\rm 5}, Yuan Tian\textsuperscript{\rm 1,2$^{*}$}, Yi Chang\textsuperscript{\rm 1,2,3\thanks{~~Joint Corresponding Author}
	}\\ 
	\textsuperscript{\rm 1}School of Artificial Intelligence, Jilin University\\
	\textsuperscript{\rm 2}Key Laboratory of Symbolic Computation and Knowledge Engineering, Jilin University\\
	\textsuperscript{\rm 3}International Center of Future Science, Jilin University\\
	\textsuperscript{\rm 4}Shenzhen Zhuiyi Technology Co., Ltd.\\ 
	\textsuperscript{\rm 5}School of Information and Library Science, University of North Carolina at Chapel Hill\\ 
	weizp19@mails.jlu.edu.cn,
	bojonesu@wezhuiyi.com,
	wangyue@email.unc.edu,\\
	yuantian@jlu.edu.cn,
	yichang@jlu.edu.cn\\
}
\date{}

\begin{document}
	
	\maketitle
	
	\input{tex/abstract}
	
	\input{tex/introduction}

	\input{tex/related_work}
	
	\input{tex/framework}
	
	\input{tex/experiments}
	
	\input{tex/conclusion.tex}
	
	\section*{Acknowledgments}
	The authors would like to thank the anonymous referees for their valuable comments. This work is supported by the National Natural Science Foundation of China (No.61976102, No.U19A2065). 
	
	\bibliography{CasRel.bib}
	\bibliographystyle{acl_natbib}
	
	\input{tex/appendix.tex}
\end{document}

%% file: tex/abstract.tex
\begin{abstract}
Extracting relational triples from unstructured text is crucial for large-scale knowledge graph construction. However, few existing works excel in solving the overlapping triple problem where multiple relational triples in the same sentence share the same entities. In this work, we introduce a fresh perspective to revisit the relational triple extraction task and propose a novel cascade binary tagging framework (\textsc{CasRel}) derived from a principled problem formulation. Instead of treating relations as discrete labels as in previous works, our new framework models relations as functions that map subjects to objects in a sentence, which naturally handles the overlapping problem. Experiments show that the \textsc{CasRel} framework already outperforms state-of-the-art methods even when its encoder module uses a randomly initialized BERT encoder, showing the power of the new tagging framework. It enjoys further performance boost when employing a pre-trained BERT encoder, outperforming the strongest baseline by 17.5 and 30.2 absolute gain in F1-score on two public datasets NYT and WebNLG, respectively. In-depth analysis on different scenarios of overlapping triples shows that the method delivers consistent performance gain across all these scenarios. The source code and data are released 
online\footnote{\url{https://github.com/weizhepei/CasRel}}.
\end{abstract}

%% file: tex/introduction.tex
\section{Introduction}
The key ingredient of a knowledge graph is relational facts, most of which consist of two entities connected by a semantic relation. These facts are in the form of ({subject, relation, object}), or $(s,r,o)$, referred to as {relational triples}. Extracting relational triples from natural language text is a crucial step towards constructing large-scale knowledge graphs.

\par Early works in relational triple extraction took a pipeline approach~\citep{zelenko2003Kernel,zhou2005Exploring,chan2011Exploiting}. It first recognizes all entities in a sentence and then performs relation classification for each entity pair. Such an approach tends to suffer from the error propagation problem since errors in early stages cannot be corrected in later stages. To tackle this problem, subsequent works proposed joint learning of entities and relations, among them are feature-based models~\citep{yu2010Jointly,li2014Incremental,miwa2014Modeling,ren2017Cotype} and, more recently, neural network-based models~\citep{gupta2016Table,katiyar2017Going,zheng2017Joint,zeng2018Extracting,fu2019GraphRel}. By replacing manually constructed features with learned representations, neural network-based models have achieved considerable success in the triple extraction task.

\begin{figure} [!t]
	\centering
	\includegraphics[width=1\linewidth]{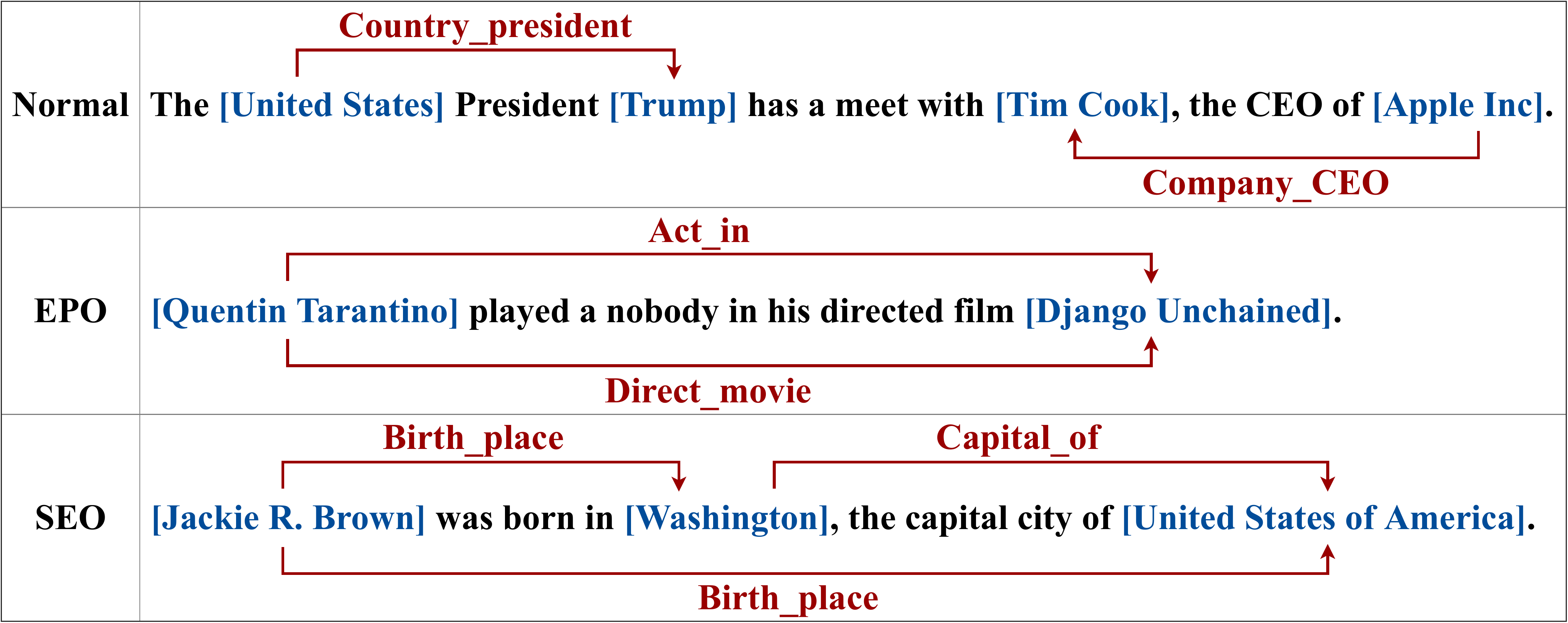}
	\caption{Examples of \emph{Normal}, \emph{EntityPairOverlap (EPO)} and \emph{SingleEntityOverlap (SEO)} overlapping patterns.}
	\label{fig:categories}
\end{figure}

\par However, most existing approaches cannot efficiently handle scenarios in which a sentence contains multiple relational triples that overlap with each other. Figure~\ref{fig:categories} illustrates these scenarios, where triples share one or two entities in a sentence. This \emph{overlapping triple problem} directly challenges conventional sequence tagging schemes that assume each token bears only one tag~\citep{zheng2017Joint}. It also brings significant difficulty to relation classification approaches where an entity pair is assumed to hold at most one relation~\citep{miwa2016End}. \citet{zeng2018Extracting} is among the first to consider the overlapping triple problem in relational triple extraction. They introduced the categories for different overlapping patterns as shown in Figure~\ref{fig:categories} and proposed a sequence-to-sequence (Seq2Seq) model with copy mechanism to extract triples. Based on the Seq2Seq model, they further investigate the impact of extraction order~\cite{zeng2019Learning} and gain considerable improvement with reinforcement learning. \citet{fu2019GraphRel} also studied the overlapping triple problem by modeling text as relational graphs with a graph convolutional networks (GCNs) based model.

\par Despite their success, previous works on extracting overlapping triples still leave much to be desired. Specifically, they all treat relations as discrete labels to be assigned to entity pairs. This formulation makes relation classification a hard machine learning problem. First, the class distribution is highly imbalanced. Among all pairs of extracted entities, most do not form valid relations, generating too many negative examples. Second, the classifier can be confused when the same entity participates in multiple valid relations (overlapping triples). Without enough training examples, the classifier can hardly tell which relation the entity participates in. As a result, the extracted triples are usually incomplete and inaccurate.

\par In this work, we start with a principled formulation of relational triple extraction {right at the triple level}. This gives rise to a general algorithmic framework that handles the overlapping triple problem {by design}. At the core of the framework is the fresh perspective that instead of treating relations as discrete labels on entity pairs, we can model relations as functions that map subjects to objects. More precisely, instead of learning relation classifiers $f(s, o) \rightarrow r$, we learn \emph{relation-specific taggers} $f_{r}(s) \rightarrow o$, each of which recognizes the possible object(s) of a given subject under a specific relation; or returns no object, indicating that there is no triple with the given subject and relation. Under this framework, triple extraction is a two-step process: first we identify all possible subjects in a sentence; then for each subject, we apply relation-specific taggers to simultaneously identify all possible relations and the corresponding objects.

\par We implement the above idea in \textsc{CasRel}, an end-to-end cascade binary tagging framework. It consists of a BERT-based encoder module, a subject tagging module, and a relation-specific object tagging module. Empirical experiments show that the proposed framework outperforms state-of-the-art methods by a large margin even when the BERT encoder is \emph{not} pre-trained, showing the superiority of the new framework itself. The framework enjoys a further large performance gain after adopting a pre-trained BERT encoder, showing the importance of rich prior knowledge in triple extraction task.

\par This work has the following main contributions:
\begin{itemize}
	\item [1.] We introduce a fresh perspective to revisit the relational triple extraction task with a principled problem formulation, which implies a general algorithmic framework that addresses the overlapping triple problem by design.
	\item [2.] We instantiate the above framework as a novel cascade binary tagging model on top of a Transformer encoder. This allows the model to combine the power of the novel tagging framework with the prior knowledge in pre-trained large-scale language models.
	\item [3.] Extensive experiments on two public datasets show that the proposed framework overwhelmingly outperforms state-of-the-art methods, achieving 17.5 and 30.2 absolute gain in F1-score on the two datasets respectively. Detailed analyses show that our model gains consistent improvement in all scenarios.
\end{itemize}

%% file: tex/related_work.tex
\section{Related Work}
Extracting relational triples from unstructured natural language texts is a well-studied task in information extraction (IE). It is also an important step for the construction of large scale knowledge graph (KG) such as DBpedia~\citep{auer2007DBpedia}, Freebase~\citep{bollacker2008Freebase} and Knowledge Vault~\citep{dong2014Knowledge}.

\par Early works~\citep{mintz2009Distant,gormley2015Improved} address the task in a pipelined manner. They extract relational triples in two separate steps: 1)~first run named entity recognition (NER) on the input sentence to identify all entities and 2)~then run relation classification (RC) on pairs of extracted entities. The pipelined methods usually suffer from the error propagation problem and neglect the relevance between the two steps.
To ease these issues, many joint models that aim to learn entities and relations jointly have been proposed. Traditional joint models~\citep{yu2010Jointly,li2014Incremental,miwa2014Modeling,ren2017Cotype} are feature-based, which heavily rely on feature engineering and require intensive manual efforts. To reduce manual work, recent studies have investigated neural network-based methods, which deliver state-of-the-art performance. However, most existing neural models like~\citep{miwa2016End} achieve joint learning of entities and relations only through parameter sharing but not joint decoding. To obtain relational triples, they still have to pipeline the detected entity pairs to a relation classifier for identifying the relation of entities. The separated decoding setting leads to a separated training objective for entity and relation, which brings a drawback that the triple-level dependencies between predicted entities and relations cannot be fully exploited. Different from those works, \citet{zheng2017Joint} achieves joint decoding by introducing a unified tagging scheme and convert the task of relational triple extraction to an end-to-end sequence tagging problem without need of NER or RC. The proposed method can directly model relational triples as a whole at the triple level since the information of entities and relations is integrated into the unified tagging scheme.

\par Though joint models (with or without joint decoding) have been well studied, most previous works ignore the problem of overlapping relational triples. \citet{zeng2018Extracting} introduced three patterns of overlapping triples and try to address the problem via a sequence-to-sequence model with copy mechanism. Recently, ~\citet{fu2019GraphRel} also study the problem and propose a graph convolutional networks (GCNs) based method. Despite their initial success, both methods still treat the relations as discrete labels of entity pairs, making it quite hard for the model to learn overlapping triples.

\par Our framework is based on a training objective that is carefully designed to directly model the relational triples as a whole like~\cite{zheng2017Joint}, i.e., to learn both entities and relations through joint decoding. Moreover, we model the relations as functions that map subjects to objects, which makes it crucially different from previous works.

%% file: tex/framework.tex
\section{The \textsc{CasRel} Framework}
The goal of relational triple extraction is to identify all possible ({subject}, {relation}, {object}) triples in a sentence, where some triples may share the same entities as subjects or objects. Towards this goal, we directly model the triples and design a training objective right at the triple level. This is in contrast to previous approaches like~\citep{fu2019GraphRel} where the training objective is defined separately for entities and relations without explicitly modeling their integration at the triple level.

\par Formally, given annotated sentence $x_j$ from the training set $D$ and a set of potentially overlapping triples $T_j = \{(s,r,o)\}$ in $x_j$, we aim to maximize the data likelihood of the training set $D$:

\begin{small}
\begin{align}
& \ \ \ \ \prod_{j=1}^{|D|} \left[ \prod_{(s, r, o)\in T_j } p((s, r, o)| x_j) \right] \label{eqn:data-likelihood} \\
& =  \prod_{j=1}^{|D|} \left[ \prod_{s\in T_j } p(s| x_j) \prod_{(r,o) \in T_j|s} p((r,o)|s,  x_j) \right] \label{eqn:obj-chain-rule} \\
& =  \prod_{j=1}^{|D|} \left[ \prod_{s\in T_j } p(s| x_j) \hspace{-0.05in} \prod_{r\in T_j|s} \hspace{-0.05in} p_r(o|s,  x_j) \hspace{-0.1in} \prod_{r \in R \backslash T_j|s} \hspace{-0.1in} p_r(o_\varnothing|s, x_j) \right] . \label{eqn:obj-relation-specific-prob}
\end{align}
\end{small}Here we slightly abuse the notation $T_j$. $s \in T_j$ denotes a subject appearing in the triples in $T_j$. $T_j|s$ is the set of triples led by subject $s$ in $T_j$. $(r,o) \in T_j|s$ is a $(r,o)$ pair in the triples led by subject $s$ in $T_j$. $R$ is the set of all possible relations. $R \backslash T_j|s$ denotes all relations except those led by $s$ in $T_j$. $o_\varnothing$ denotes a ``null'' object (explained below). 

\par Eq.~\eqref{eqn:obj-chain-rule} applies the chain rule of probability.
Eq.~\eqref{eqn:obj-relation-specific-prob} exploits the crucial fact that for a given subject $s$, any relation relevant to $s$ (those in $T_j|s$) would lead to corresponding objects in the sentence, and all other relations would necessarily have no object in the sentence, {i.e.} a ``null'' object.

\par This formulation provides several benefits. First, since the data likelihood starts at the triple level, optimizing this likelihood corresponds to directly optimizing the final evaluation criteria at the triple level. Second, by making no assumption on how multiple triples may share entities in a sentence, it handles the overlapping triple problem \emph{by design}. Third, the decomposition in Eq.~$\eqref{eqn:obj-relation-specific-prob}$ inspires a novel tagging scheme for triple extraction: we learn a subject tagger $p(s| x_j)$ that recognizes subject entities in a sentence; and for each relation $r$, we learn an object tagger $p_r(o|s,  x_j)$  that recognizes relation-specific objects for a given subject. In this way we can model each relation as a function that maps subjects to objects, as opposed to classifying relations for (\emph{subject}, \emph{object}) pairs.

\par Indeed, this novel tagging scheme allows us to extract multiple triples at once: we first run the subject tagger to find all possible subjects in the sentence, and then for each subject found, apply relation-specific object taggers to find all relevant relations and the corresponding objects.

\par The key components in the above general framework, i.e., the subject tagger and relation-specific object taggers, can be instantiated in many ways. In this paper, we instantiate them as binary taggers on top of a deep bidirectional Transformer BERT~\citep{devlin2019BERT}. We describe its detail below.

	\begin{figure*} [!t]
		\centering
		\includegraphics[width=1\linewidth]{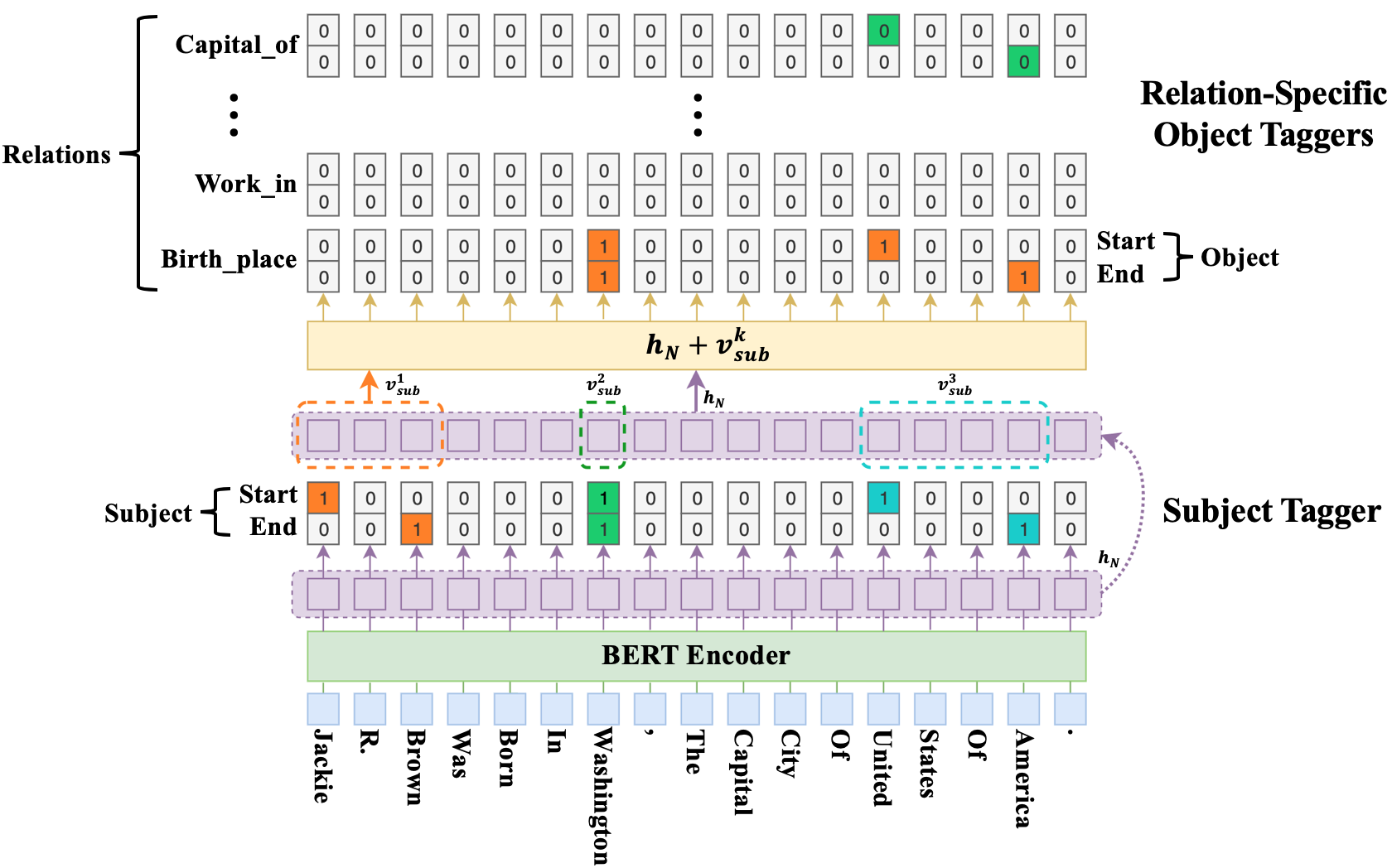} 
		\caption{An overview of the proposed \textsc{CasRel} framework. In this example, there are three candidate subjects detected at the low level, while the presented 0/1 tags at high level are specific to the first subject \emph{Jackie R. Brown}, i.e., a snapshot of the iteration state when $k = 1$ is shown as above. For the subsequent iterations ($k = 2, 3$), the results at high level will change, reflecting different triples detected. For instance, when $k = 2$, the high-level orange (green) blocks will change to 0 (1), respectively, reflecting the relational triple \emph{(Washington, Capital\_of, United States Of America)} led by the second candidate subject \emph{Washington}.}
		\label{fig:model}
	\end{figure*}

\subsection{BERT Encoder}
The encoder module extracts feature information $\mathbf x_j$  from sentence $x_j$, which will feed into subsequent tagging modules\footnote{This paper uses boldface letters to denote vectors and matrices.}. We employ a pre-trained BERT model~\cite{devlin2019BERT} to encode the context information.

\par Here we briefly review BERT, a multi-layer bidirectional Transformer based language representation model. It is designed to learn deep representations by jointly conditioning on both left and right context of each word, and it has recently been proven surprisingly effective in many downstream tasks~\cite{zhong2019Knowledge}. Specifically, it is composed of a stack of $N$ identical Transformer blocks. We denote the Transformer block as $Trans(\mathbf x)$, in which $\mathbf x$ represents the input vector. The detailed operations are as follows:
\begin{align}
	\mathbf h_0 = \  & \mathbf S \mathbf W_s + \mathbf W_p\label{func:1}\\
	\mathbf h_\alpha = \  & Trans(\mathbf h_{\alpha-1}), \alpha\in [1, N]\label{func:2}
\end{align}
where $\mathbf S$ is the matrix of one-hot vectors of sub-words\footnote{We use WordPiece embeddings~\citep{wu2016Googles} to represent words in vector space as in BERT~\cite{devlin2019BERT}. Each word in the input sentence will be tokenized to fine-grained tokens, \emph{i.e.}, sub-words.} indices in the input sentence, $\mathbf W_s$ is the sub-words embedding matrix, $\mathbf W_p$ is the positional embedding matrix where $p$ represents the position index in the input sequence, $\mathbf h_\alpha$ is the hidden state vector, i.e., the context representation of input sentence at $\alpha$-th layer and $N$ is the number of Transformer blocks. Note that in our work the input is a single text sentence instead of sentence pair, hence the segmentation embedding as described in original BERT paper was not taken into account in Eq.~\eqref{func:1}. For a more comprehensive description of the Transformer structure, we refer readers to~\citep{vaswani2017Attention}.

\subsection{Cascade Decoder}
Now we describe our instantiation of the novel cascade binary tagging scheme inspired by the previous formulation. The basic idea is to extract triples in two cascade steps. First, we detect subjects from the input sentence. Then for each candidate subject, we check all possible relations to see if a relation can associate objects in the sentence with that subject. Corresponding to the two steps, the cascade decoder consists of two modules as illustrated in Figure \ref{fig:model}: a subject tagger; and a set of relation-specific object taggers.

\paragraph{Subject Tagger}
The low level tagging module is designed to recognize all possible subjects in the input sentence by directly decoding the encoded vector $\mathbf h_N$ produced by the $N$-layer BERT encoder. More precisely, it adopts two identical binary classifiers to detect the start and end position of subjects respectively by assigning each token a binary tag (0/1) that indicates whether the current token corresponds to a start or end position of a subject. The detailed operations of the subject tagger on each token are as follows:
\begin{align}
		p_i^{start\_s} = \sigma(\mathbf W_{start}\mathbf x_i+\mathbf b_{start})\label{func:low1}\\
	p_i^{end\_s} = \sigma(\mathbf W_{end}\mathbf x_i+\mathbf b_{end})\label{func:low2}
\end{align}
where $p_i^{start\_s}$ and $p_i^{end\_s}$ represent the probability of identifying the $i$-th token in the input sequence as the start and end position of a subject, respectively. The corresponding token will be assigned with a tag 1 if the probability exceeds a certain threshold or with a tag 0 otherwise. $\mathbf x_i$ is the encoded representation of the $i$-th token in the input sequence, \emph{i.e.}, $\mathbf x_i$ = $\mathbf h_N[i]$, where $\mathbf W_{(\cdot)}$ represents the trainable weight, and $\mathbf b_{(\cdot)}$ is the bias and $\sigma$ is the sigmoid activation function.

\par The subject tagger optimizes the following likelihood function to identify the span of subject $s$ given a sentence representation $\mathbf x$:
\begin{align}
& p_{\theta}(s |\mathbf x) \nonumber \\ = & \prod_{t \in \{start\_s,end\_s\}} \prod_{i=1}^{L}  \left(p_i^t\right)^{\mathbf I\{y_i^t=1\}}
\left(1-p_i^t\right)^{\mathbf I\{y_i^t=0\}} \ . \label{eqn:subject-tagging}
\end{align}
where $L$ is the length of the sentence. $\mathbf I\{z\} = 1$ if $z$ is true and 0 otherwise. $y_{i}^{start\_s}$ is the binary tag of subject start position for the $i$-th token in $\mathbf{x}$, and $y_{i}^{end\_s}$ indicates the subject end position. The parameters $\theta = \{ \mathbf W_{start}, \mathbf b_{start}, \mathbf W_{end}, \mathbf b_{end} \}$.

\par For multiple subjects detection, we adopt the nearest start-end pair match principle to decide the span of any subject based on the results of the start and end position taggers. For example, as shown in Figure~\ref{fig:model}, the nearest end token to the first start token ``Jackie" is ``Brown", hence the detected result of the first subject span will be ``Jackie R. Brown". Notably, to match an end token for a given start token, we don't consider tokens whose position is prior to the position of the given token. Such match strategy is able to maintain the integrity of any entity span if the start and end positions are both correctly detected due to the natural continuity of any entity span in a given sentence.

\paragraph{Relation-specific Object Taggers} 
The high level tagging module simultaneously identifies the objects as well the involved relations with respect to the subjects obtained at lower level. As Figure~\ref{fig:model} shows, it consists of a set of relation-specific object taggers with the same structure as subject tagger in low level module for all possible relations. All object taggers will identify the corresponding object(s) for each detected subject at the same time. Different from subject tagger directly decoding the encoded vector $\mathbf h_N$, the relation-specific object tagger takes the subject features into account as well. The detailed operations of the relation-specific object tagger on each token are as follows:
\begin{align}
p_i^{start\_o} = \sigma(\mathbf W^{r}_{start}(\mathbf x_i + \mathbf v_{sub}^k)+\mathbf b^{r}_{start})\label{func:high1} \\
p_i^{end\_o} = \sigma(\mathbf W^{r}_{end}(\mathbf x_i + \mathbf v_{sub}^k)+\mathbf b^{r}_{end})\label{func:high2}
\end{align}
where $p_i^{start\_o}$ and $p_i^{end\_o}$ represent the probability of identifying the $i$-th token in the input sequence as the start and end position of a object respectively, and $\mathbf v^{k}_{sub}$ represents the encoded representation vector of the $k$-th subject detected in low level module.

\par For each subject, we iteratively apply the same decoding process on it. Note that the subject is usually composed of multiple tokens, to make the additions of $\mathbf x_i$ and $\mathbf v_{sub}^k$ in Eq.~\eqref{func:high1} and Eq.~\eqref{func:high2} possible, we need to keep the dimension of two vectors consistent. To do so, we take the averaged vector representation between the start and end tokens of the $k$-th subject as $\mathbf v_{sub}^k$.

\par The object tagger for relation $r$ optimizes the following likelihood function to identify the span of object $o$ given a sentence representation $\mathbf x$ and a subject $s$:
	\begin{align}
	& p_{\phi_r}(o |s,\mathbf x) \nonumber \\ 
	= & \prod_{t \in \{start\_o,end\_o\}} \prod_{i=1}^{L}  \left(p_i^t\right)^{\mathbf I\{y_i^t=1\}}
	\left(1-p_i^t\right)^{\mathbf I\{y_i^t=0\}} \ . \label{eqn:object-tagging}
	\end{align}
where $y_{i}^{start\_o}$ is the binary tag of object start position for the $i$-th token in $\mathbf{x}$, and $y_{i}^{end\_o}$ is the tag of object end position for the $i$-th token. For a ``null'' object $o_{\varnothing}$, the tags $y_{i}^{start\_{o_{\varnothing}}} = y_{i}^{end\_{o_{\varnothing}}} = 0$ for all $i$. The parameters $\phi_r = \{ \mathbf W_{start}^r, \mathbf b_{start}^r, \mathbf W_{end}^r, \mathbf b_{end}^r \}$. 

\par Note that in the high level tagging module, the relation is also decided by the output of object taggers. For example, the relation ``Work\_in'' does not hold between the detected subject ``Jackie R. Brown" and the candidate object ``Washington''. Therefore, the object tagger for relation ``Work\_in" will not identify the span of ``Washington", i.e., the output of both start and end position are all zeros as shown in Figure~\ref{fig:model}. In contrast, the relation ``Birth\_place" holds between ``Jackie R. Brown" and ``Washington", so the corresponding object tagger outputs the span of the candidate object ``Washington". In this setting, the high level module is capable of simultaneously identifying the relations and objects with regard to the subjects detected in low level module.
\subsection{Data Log-likelihood Objective} 
\par Taking log of Eq.~\eqref{eqn:obj-relation-specific-prob}, the objective $J(\Theta)$ is:
\begin{align}
 \sum_{j=1}^{|D|} &\left[ \sum_{s\in T_j}\log p_\theta(s|\mathbf x_j) + \sum_{r\in T_j|s} \log p_{\phi_r}(o|s,\mathbf x_j) \right. \nonumber \\
&  + \left. \sum_{r \in R \backslash T_j|s} \log p_{\phi_r}(o_{\varnothing}|s,\mathbf x_j) \right] \ .
\end{align}
where parameters $\Theta = \{\theta, \{\phi_r\}_{r\in R}\}$. $p_\theta(s|\mathbf x)$ is defined in Eq.~\eqref{eqn:subject-tagging} and $p_{\phi_r}(o|s,\mathbf x)$ is defined in Eq.~\eqref{eqn:object-tagging}. We train the model by maximizing $J(\Theta)$ through Adam stochastic gradient descent~\citep{kingma2014Adam} over shuffled mini-batches.

%% file: tex/experiments.tex
\section{Experiments} 
	\subsection{Experimental Setting}
	\renewcommand{\arraystretch}{1} 
	\begin{table}[!t] 
		\centering 
		\small
		\begin{tabular}{ccccc} 
			\toprule
			\multirow{2}{*}{Category}& 
			\multicolumn{2}{c}{NYT}&\multicolumn{2}{c}{WebNLG}\\
			\cmidrule(lr){2-3} \cmidrule(lr){4-5} &Train&Test&Train&Test\\
			\midrule 
			\textit{Normal}&37013&3266&1596&246\\
			\textit{EPO}&9782&978&227&26\\
			\textit{SEO}&14735&1297&3406&457\\
			\midrule 
			ALL&56195&5000&5019&703\\
			\bottomrule 
		\end{tabular} 
		\caption{Statistics of datasets. Note that a sentence can belong to both \emph{EPO} class and \emph{SEO} class.} 
		\label{tab:statistics} 
	\end{table}
	
	\renewcommand{\arraystretch}{1} 
	\begin{table*}[!tp] 
		\centering
		\small
		\begin{tabular}{lllllll} 
			\toprule
			\multirow{2}{*}{Method}& 
			\multicolumn{3}{c}{NYT}&\multicolumn{3}{c}{WebNLG}\\
			\cmidrule(lr){2-4} \cmidrule(lr){5-7} &\textit{Prec.}&\textit{Rec.}&\textit{F1}&\textit{Prec.}&\textit{Rec.}&\textit{F1}\\
			\midrule 
			NovelTagging
			~\cite{zheng2017Joint}
			&62.4&31.7&42.0&52.5&19.3&28.3\\ 
			CopyR$_{OneDecoder}$
			~\cite{zeng2018Extracting}
			&59.4&53.1&56.0&32.2&28.9&30.5\\
			CopyR$_{MultiDecoder}$
			~\cite{zeng2018Extracting}
			&61.0&56.6&58.7&37.7&36.4&37.1\\
			GraphRel$_{1p}$
			~\cite{fu2019GraphRel}
			&62.9&57.3&60.0&42.3&39.2&40.7\\
			GraphRel$_{2p}$
			~\cite{fu2019GraphRel}
			&63.9&60.0&61.9&44.7&41.1&42.9\\
			CopyR$_{RL}$
			~\cite{zeng2019Learning}
			&77.9&67.2&72.1&63.3&59.9&61.6\\
			CopyR$_{RL}^*$
			&72.8&69.4&71.1&60.9&61.1&61.0\\
			\midrule
			\textsc{CasRel}$_{random}$&81.5&75.7&78.5&84.7&79.5&82.0\\
			\textsc{CasRel}$_{LSTM}$&84.2&83.0&83.6&86.9&80.6&83.7\\
			\textsc{CasRel}&\textbf{89.7}& \textbf{89.5} &\textbf{89.6}&\textbf{93.4} & \textbf{90.1}&\textbf{91.8}\\
			\bottomrule 
		\end{tabular} 
		\caption{Results of different methods on NYT and WebNLG datasets. Our re-implementation is marked by *.}
		\label{tab:result} 
	\end{table*}

\paragraph{Datasets and Evaluation Metrics}
We evaluate the framework on two public datasets NYT~\citep{riedel2010Modeling} and WebNLG~\citep{gardent2017Creating}. NYT dataset was originally produced by distant supervision method. It consists of 1.18M sentences with 24 predefined relation types. WebNLG dataset was originally created for Natural Language Generation (NLG) tasks and adapted by~\citep{zeng2018Extracting} for relational triple extraction task. It contains 246 predefined relation types. The sentences in both datasets commonly contain multiple relational triples, thus NYT and WebNLG datasets suit very well to be the testbed for evaluating model on extracting overlapping relational triples\footnote{Datasets such as ACE, Wiki-KBP have few overlapping triples in the sentences hence are not suitable for evaluating the performance of overlapping triple extraction. Nonetheless, to validate the generality of the proposed framework, we also conduct supplemental experiments on these datasets along with the comparison of 12 recent strong baselines. The results of the comprehensive comparison, which show consistent superiority of our model over most compared methods, can be found in Appendix~\ref{appendix:sup_exp}.}. We use the datasets released by~\citep{zeng2018Extracting}, in which NYT contains 56195 sentences for training, 5000 sentences for validation, and 5000 sentences for test, and WebNLG contains 5019 sentences for training, 500 sentences for validation and 703 sentences for test. According to different overlapping patterns of relational triples, we split the sentences into three categories, namely, \emph{Normal}, \emph{EntityPairOverlap (EPO)} and \emph{SingleEntityOverlap (SEO)} for detailed experiments on different types of overlapping relational triples. The statistics of the two datasets are described in Table~\ref{tab:statistics}.

\par Following previous work~\citep{fu2019GraphRel}, an extracted relational triple \emph{(subject, relation, object)} is regarded as correct only if the relation and the heads of both subject and object are all correct. For fair comparison, we report the standard micro Precision (Prec.), Recall (Rec.) and F1-score as in line with baselines.\label{exp:metric}

\paragraph{Implementation Details}
The hyper-parameters are determined on the validation set. More implementation details are described in Appendix~\ref{appendix:exp_detail}.

\subsection{Experimental Result}
\paragraph{Compared Methods}
We compare our model with several strong state-of-the-art models, namely, \textbf{NovelTagging}~\cite{zheng2017Joint}, \textbf{CopyR}~\cite{zeng2018Extracting}, \textbf{GraphRel}~\cite{fu2019GraphRel} and \textbf{CopyR$_{RL}$}~\cite{zeng2019Learning}. The reported results for the above baselines are directly copied from the original published literature.

\par Note that we instantiate the \textsc{CasRel} framework on top of a pre-trained BERT model to combine the power of the proposed novel tagging scheme and the pre-learned prior knowledge for better performance. To evaluate the impact of introducing the Transformer-based BERT model, we conduct a set of ablation tests.
\textbf{\textsc{CasRel}}$_{random}$ is the framework where all parameters of BERT are randomly initialized; 
\textbf{\textsc{CasRel}}$_{LSTM}$ is the framework instantiated on a LSTM-based structure as in~\citep{zheng2017Joint} with pre-trained Glove embedding~\citep{pennington2014Glove}; 
\textbf{\textsc{CasRel}} is the full-fledged framework using pre-trained BERT weights. 

\begin{figure*} [!t]
	\centering
	\subfigure[F1 of Normal Class]{
		\includegraphics[width=0.315\linewidth]{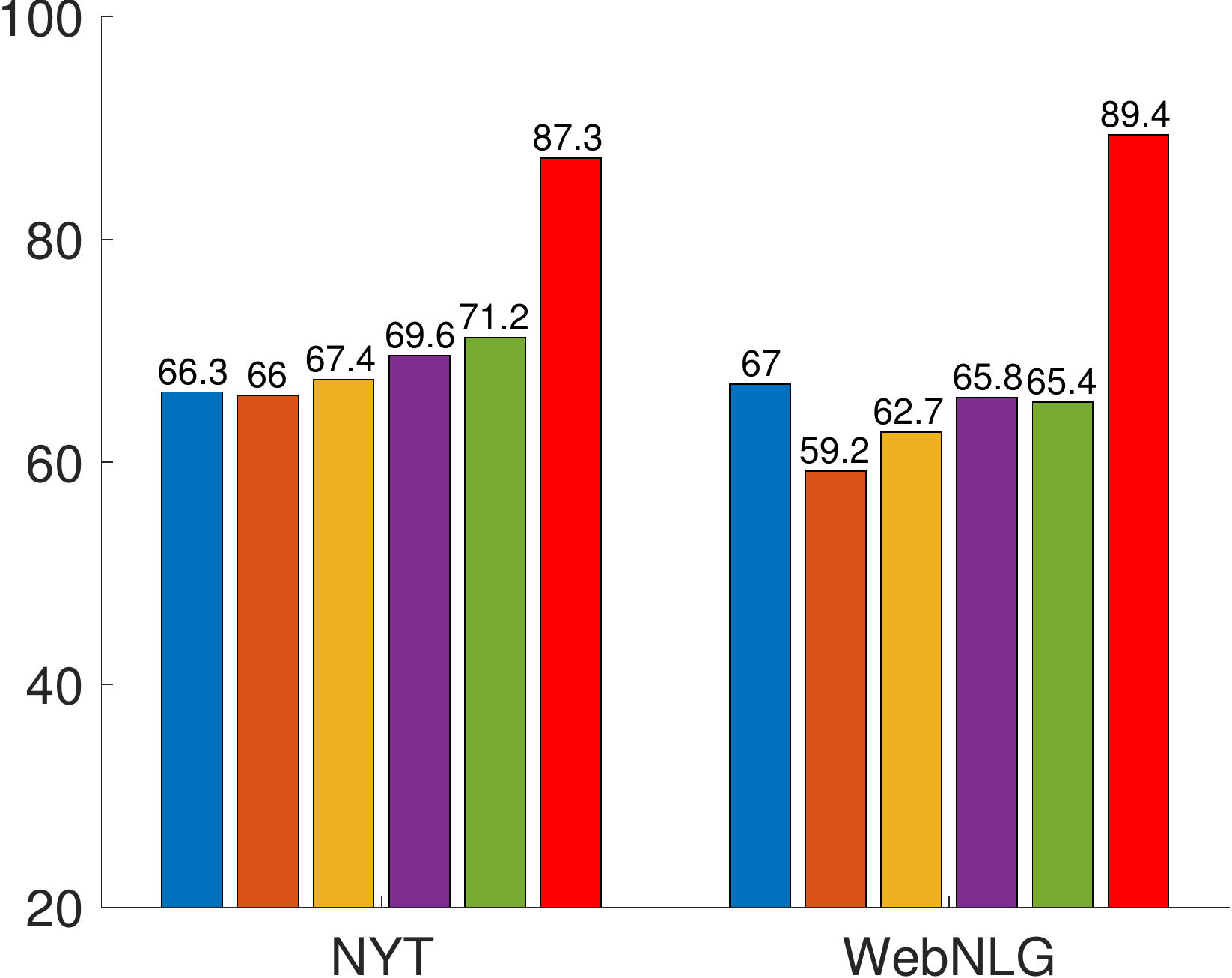}
	}
	\subfigure[F1 of EntityPairOverlap Class]{
		\includegraphics[width=0.315\linewidth]{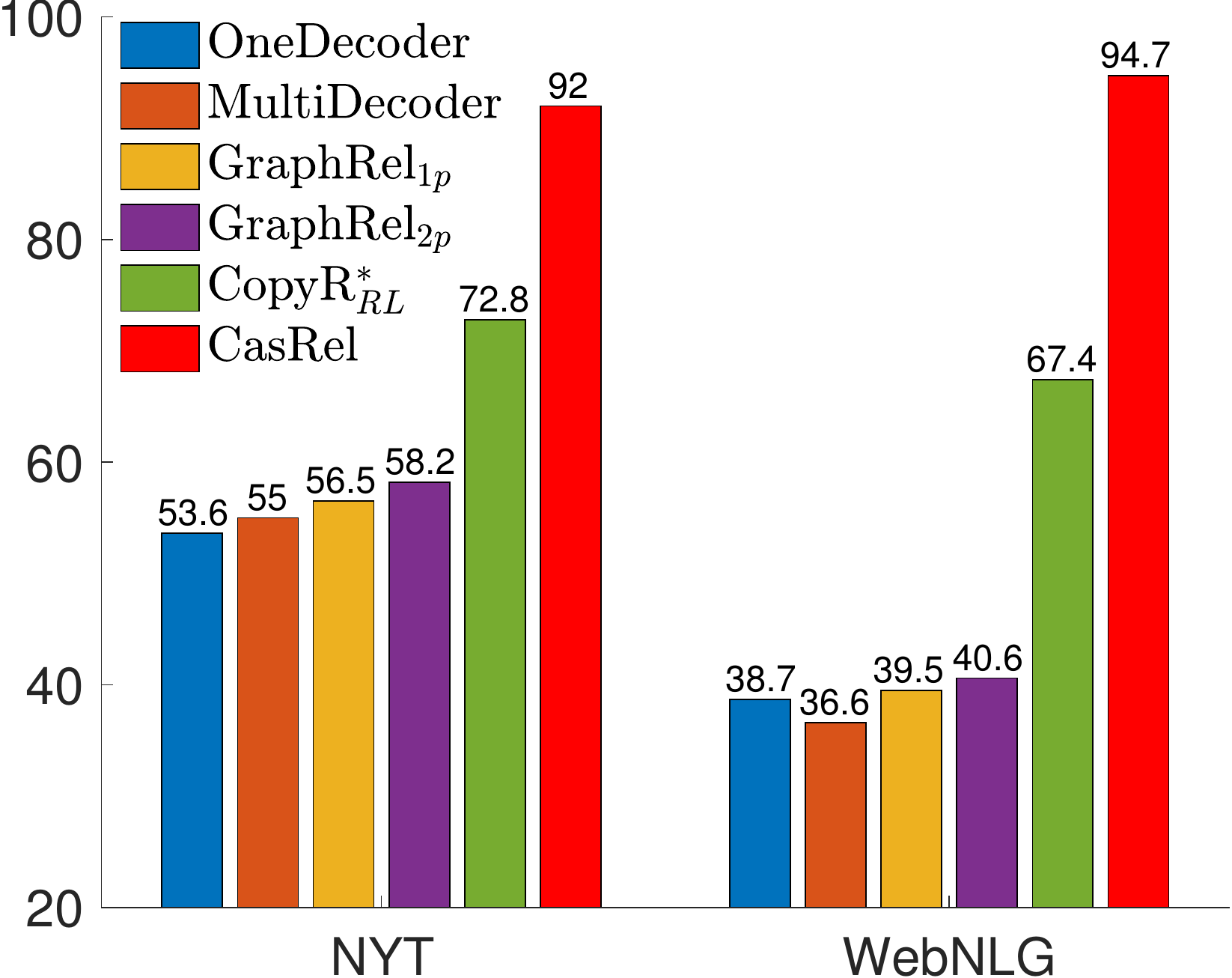}
	}
	\subfigure[F1 of SingleEntityOverlap Class]{
		\includegraphics[width=0.315\linewidth]{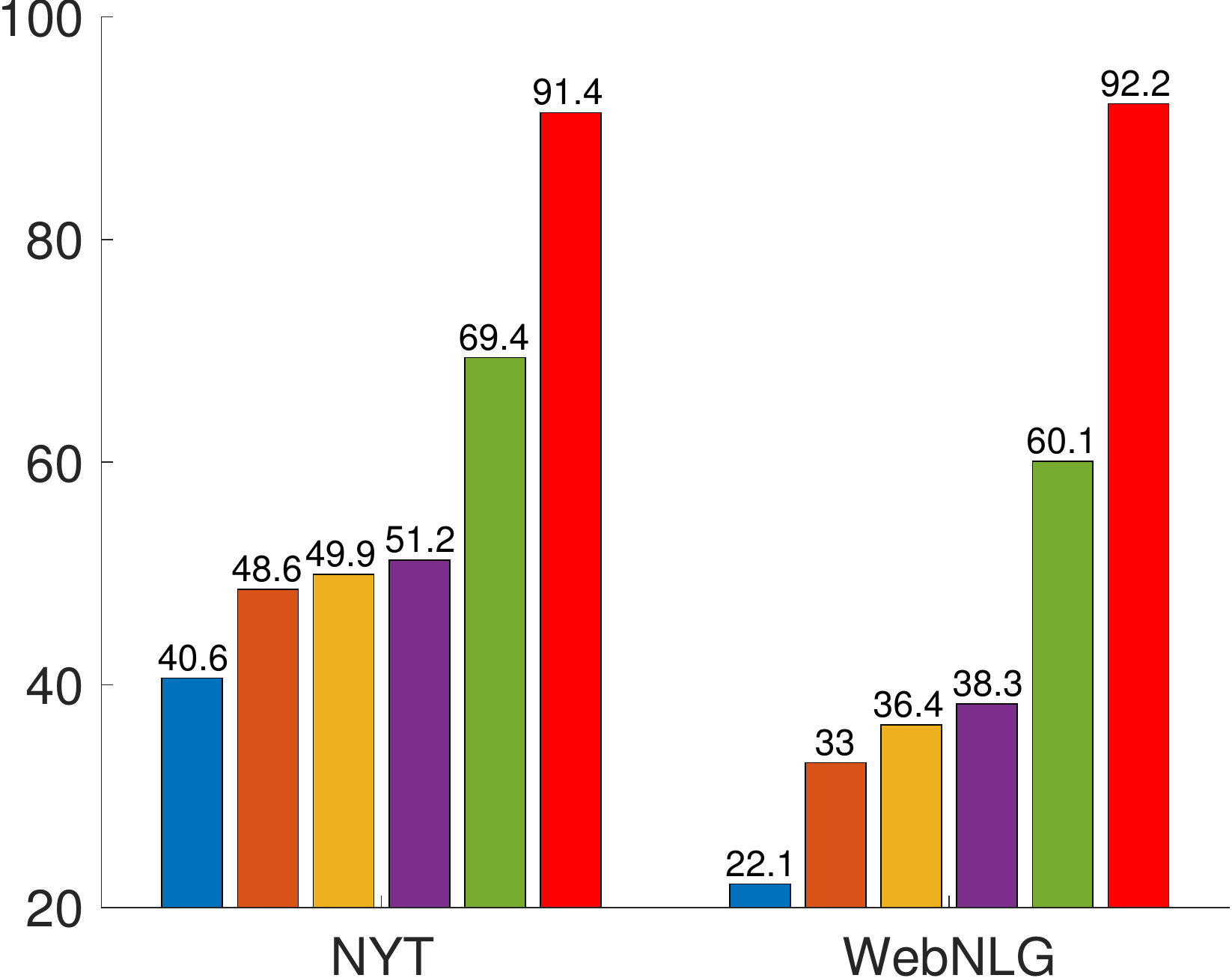}
	}
	\caption{F1-score of extracting relational triples from sentences with different overlapping pattern.}
	\label{fig:result_type}
\end{figure*}

\renewcommand{\arraystretch}{1} 
\begin{table*}[!tp] 
	\centering
	\small
	\begin{tabular}{lllllllllll} 
		\toprule
		\multirow{2}{*}{Method}& 
		\multicolumn{5}{c}{NYT}&\multicolumn{5}{c}{WebNLG}\\
		\cmidrule(lr){2-6} \cmidrule(lr){7-11} &\textit{N=1}&\textit{N=2}&\textit{N=3}&\textit{N=4}&\textit{N$\geq$5}
		&\textit{N=1}&\textit{N=2}&\textit{N=3}&\textit{N=4}&\textit{N$\geq$5}\\
		\midrule 
		CopyR$_{OneDecoder}$
		&66.6&52.6&49.7&48.7&20.3&65.2&33.0&22.2&14.2&13.2\\ 
		CopyR$_{MultiDecoder}$
		&67.1&58.6&52.0&53.6&30.0&59.2&42.5&31.7&24.2&30.0\\
		GraphRel$_{1p}$
		&69.1&59.5&54.4&53.9&37.5&63.8&46.3&34.7&30.8&29.4\\
		GraphRel$_{2p}$
		&71.0&61.5&57.4&55.1&41.1&66.0&48.3&37.0&32.1&32.1\\
		CopyR$_{RL}^*$
		&71.7&72.6&72.5&77.9&45.9&63.4&62.2&64.4&57.2&55.7\\
		\midrule
		\textsc{CasRel}&\textbf{88.2}&\textbf{90.3}&\textbf{91.9}&\textbf{94.2}& \textbf{83.7 (+37.8)}&\textbf{89.3}&\textbf{90.8}&\textbf{94.2}&\textbf{92.4}&\textbf{90.9 (+35.2)}\\
		\bottomrule 
	\end{tabular} 
	\caption{F1-score of extracting relational triples from sentences with different number (denoted as N) of triples.}
	\label{tab:result_num} 
\end{table*}

\paragraph{Main Results}
Table~\ref{tab:result} shows the results of different baselines for relational triple extraction on two datasets. The \textsc{CasRel} model overwhelmingly outperforms all the baselines in terms of all three evaluation metrics and achieves encouraging 17.5\% and 30.2\% improvements in F1-score over the best state-of-the-art method~\cite{zeng2019Learning} on NYT and WebNLG datasets respectively. Even without taking advantage of the pre-trained BERT, \textsc{CasRel}$_{random}$ and \textsc{CasRel}$_{LSTM}$ are still competitive to existing state-of-the-art models. This validates the utility of the proposed cascade decoder that adopts a novel binary tagging scheme. The performance improvements from \textsc{CasRel}$_{random}$ to \textsc{CasRel} highlight the importance of the prior knowledge in a pre-trained language model.

\par We can also observe from the table that there is a significant gap between the performance on NYT and WebNLG datasets for existing models, and we believe this gap is due to their drawbacks in dealing with overlapping triples. More precisely, as presented in Table~\ref{fig:categories}, we can find that NYT dataset is mainly comprised of \emph{Normal} class sentences while the majority of sentences in WebNLG dataset belong to \emph{EPO} and \emph{SEO} classes. Such inconsistent data distribution of two datasets leads to a comparatively better performance on NYT and a worse performance on WebNLG for all the baselines, exposing their drawbacks in extracting overlapping relational triples. In contrast, the \textsc{CasRel} model and its variants (i.e., \textsc{CasRel}$_{random}$ and \textsc{CasRel}$_{LSTM}$ ) all achieve a stable and competitive performance on both NYT and WebNLG datasets, demonstrating the effectiveness of the proposed framework in solving the overlapping problem.

\paragraph{Detailed Results on Different Types of Sentences} \label{analysis}
To further study the capability of the proposed \textsc{CasRel} framework in extracting overlapping relational triples, we conduct two extended experiments on different types of sentences and compare the performance with previous works.

\par The detailed results on three different overlapping patterns are presented in Figure~\ref{fig:result_type}. It can be seen that the performance of most baselines on \emph{Normal}, \emph{EPO} and \emph{SEO} presents a decreasing trend, reflecting the increasing difficulty of extracting relational triples from sentences with different overlapping patterns. That is, among the three overlapping patterns, \emph{Normal} class is the easiest pattern while \emph{EPO} and \emph{SEO} classes are the relatively harder ones for baseline models to extract. In contrast, the proposed \textsc{CasRel} model attains consistently strong performance over all three overlapping patterns, especially for those hard patterns. We also validate the \textsc{CasRel}'s capability in extracting relational triples from sentences with different number of triples. We split the sentences into five classes and Table~\ref{tab:result_num} shows the results. Again, the \textsc{CasRel} model achieves excellent performance over all five classes. Though it's not surprising to find that the performance of most baselines decreases with the increasing number of relational triples that a sentence contains, some patterns still can be observed from the performance changes of different models. Compared to previous works that devote to solving the overlapping problem in relational triple extraction, our model suffers the least from the increasing complexity of the input sentence. Though the \textsc{CasRel} model gain considerable improvements on all five classes compared to the best state-of-the-art method CopyR$_{RL}$~\cite{zeng2019Learning}, the greatest improvement of F1-score on the two datasets both come from the most difficult class (N$\geq$5), indicating that our model is more suitable for complicated scenarios than the baselines.

\par Both of these experiments validate the superiority of the proposed cascade binary tagging framework in extracting multiple (possibly overlapping) relational triples from complicated sentences compared to existing methods. Previous works have to explicitly predict all possible relation types contained in a given sentence, which is quite a challenging task, and thus many relations are missing in their extracted results. In contrast, our \textsc{CasRel} model side-steps the prediction of relation types and tends to extract as many relational triples as possible from a given sentence. We attribute this to the relation-specific object tagger setting in high level tagging module of the cascade decoder that considers all the relation types simultaneously.

%% file: tex/conclusion.tex
\section{Conclusion}
In this paper, we introduce a novel cascade binary tagging framework (\textsc{CasRel}) derived from a principled problem formulation for relational triple extraction. Instead of modeling relations as discrete labels of entity pairs, we model the relations as functions that map subjects to objects, which provides a fresh perspective to revisit the relational triple extraction task. As a consequent, our model can simultaneously extract multiple relational triples from sentences, without suffering from the overlapping problem. We conduct extensive experiments on two widely used datasets to validate the effectiveness of the proposed \textsc{CasRel} framework. Experimental results show that our model overwhelmingly outperforms state-of-the-art baselines over different scenarios, especially on the extraction of overlapping relational triples.

%% file: tex/appendix.tex
\newpage
\section*{Appendix}
\appendix
\section{Implementation Details} \label{appendix:exp_detail}
We adopt mini-batch mechanism to train our model with batch size as 6; the learning rate is set to 1e$^{-5}$; the hyper-parameters are determined on the validation set. We also adopt early stopping mechanism to prevent the model from over-fitting. Specifically, we stop the training process when the performance on validation set does not gain any improvement for at least 7 consecutive epochs.  The number of stacked bidirectional Transformer blocks $N$ is 12 and the size of hidden state $\mathbf h_N$ is 768. The pre-trained BERT model we used is [BERT-Base, Cased]\footnote{Available at: https://storage.googleapis.com/bert\_models\\/2018\_10\_18/cased\_L-12\_H-768\_A-12.zip},which contains 110M parameters.

\par For fair comparison, the max length of input sentence to our model is set to 100 words as previous works~\citep{zeng2018Extracting,fu2019GraphRel} suggest. We did not tune the threshold for both start and end position taggers to predict tag 1, but heuristically set the threshold to $0.5$ as default. The performance might be better after carefully tuning the threshold, however it is beyond the research scope of this paper.

\section{Error Analysis}
To explore the factors that affect the extracted relational triples of the \textsc{CasRel} model, we analyze the performance on predicting different elements of the triple \emph{(E1, R, E2)} where \emph{E1} represents the subject entity, \emph{E2} represents the object entity and \emph{R} represents the relation between them. An element like \emph{(E1, R)} is regarded as correct only if the subject and the relation in the predicted triple \emph{(E1, R, E2)} are both correct, regardless of the correctness of the predicted object. Similarly, we say an instance of \emph{E1} is correct as long as the subject in the extracted triple is correct, so is \emph{E2} and \emph{R}.

\renewcommand{\arraystretch}{1} 
\begin{table}[t!] 
	\centering 
	\small
	\resizebox{1\linewidth}{!}{
		\begin{tabular}{ccccccc} 
			\toprule
			\multirow{2}{*}{Element}& 
			\multicolumn{3}{c}{NYT}&\multicolumn{3}{c}{WebNLG}\\
			\cmidrule(lr){2-4} \cmidrule(lr){5-7} &\textit{Prec.}&\textit{Rec.}&\textit{F1}&\textit{Prec.}&\textit{Rec.}&\textit{F1}\\
			\midrule 
			\textit{E1}&94.6&92.4&93.5&98.7&92.8&95.7\\ 
			\textit{E2}&94.1&93.0&93.5&97.7&93.0&95.3\\ 
			\textit{R}&96.0&93.8&94.9&96.6&91.5&94.0\\
			\midrule 
			\textit{(E1, R)}&93.6&90.9&92.2&94.8&90.3&92.5\\ 
			\textit{(R, E2)}&93.1&91.3&92.2&95.4&91.1&93.2\\
			\textit{(E1, E2)}&89.2&90.1&89.7&95.3&91.7&93.5\\
			\midrule 
			\textit{(E1, R, E2)}&89.7&89.5&89.6&93.4&90.1&91.8\\
			\bottomrule 
		\end{tabular} 
	}
	\caption{Results on relational triple elements.} 
	\label{tab:entity-result} 
\end{table}

\par Table~\ref{tab:entity-result} shows the results on different relational triple elements. For NYT, the performance on \emph{E1} and \emph{E2} are consistent with that on \emph{(E1, R)} and \emph{(R, E2)}, demonstrating the effectiveness of our proposed framework in identifying both subject and object entity mentions. We also find that there is only a trivial gap between the F1-score on \emph{(E1, E2)} and \emph{(E1, R, E2)}, but an obvious gap between \emph{(E1, R, E2)} and \emph{(E1, R)/(R, E2)}. It reveals that most relations for the entity pairs in extracted triples are correctly identified while some extracted entities fail to form a valid relational triple. In other words, it implies that identifying relations is somehow easier than identifying entities for our model.

\par In contrast to NYT, for WebNLG, the performance gap between \emph{(E1, E2)} and \emph{(E1, R, E2)}  is comparatively larger than that between \emph{(E1, R, E2)} and \emph{(E1, R)/(R, E2)}. It shows that misidentifying the relations will bring more performance degradation than misidentifying the entities. Such observation also indicates that it's more challenging for the proposed \textsc{CasRel} model to identify the relations than entities in WebNLG, as opposed to what we observed in NYT.  We attribute such difference to the different number of relations contained in the two datasets (i.e., 24 in NYT and 246 in WebNLG), which makes the identification of relation much harder in WebNLG.

\renewcommand{\arraystretch}{1} 
\begin{table*}[!tp] 
	\centering 
	\small
	\resizebox{1\linewidth}{!}{
		\begin{tabular}{l lll lll lll lll} 
			\toprule
			\multirow{3}{*}{Method}
			&\multicolumn{9}{c}{\emph{Partial Match}}
			&\multicolumn{3}{c}{\emph{Exact Match}}\\
			\cmidrule(lr){2-10} 
			\cmidrule(lr){11-13} 
			&\multicolumn{3}{c}{ACE’04}
			&\multicolumn{3}{c}{NYT10-HRL}
			&\multicolumn{3}{c}{NYT11-HRL}
			&\multicolumn{3}{c}{Wiki-KBP}\\
			\cmidrule(lr){2-4} 
			\cmidrule(lr){5-7} 
			\cmidrule(lr){8-10}
			\cmidrule(lr){11-13}
			&\textit{Prec.}&\textit{Rec.}&\textit{F1}
			&\textit{Prec.}&\textit{Rec.}&\textit{F1}
			&\textit{Prec.}&\textit{Rec.}&\textit{F1}
			&\textit{Prec.}&\textit{Rec.}&\textit{F1}\\
			\midrule 
			~\citet{chan2011Exploiting}
			&42.9&38.9&40.8&--&--&--&--&--&--&--&--&--\\
			MultiR
			~\citep{hoffmann2011Knowledge}
			&--&--&--&--&--&--&32.8&30.6&31.7&30.1&53.0&38.0\\
			DS-Joint
			~\citep{li2014Incremental}
			&\textbf{64.7}&38.5&48.3&--&--&--&--&--&--&--&--&--\\
			FCM
			~\citep{gormley2015Improved}
			&--&--&--&--&--&--&43.2&29.4&35.0&--&--&--\\
			SPTree
			~\citep{miwa2016End}
			&--&--&--&49.2&55.7&52.2&52.2&54.1&53.1&--&--&--\\
			CoType
			~\citep{ren2017Cotype}
			&--&--&--&--&--&--&48.6&38.6&43.0&31.1&\textbf{53.7}&38.8\\ 
			~\citet{katiyar2017Going}
			&50.2&\textbf{48.8}&49.3&--&--&--&--&--&--&--&--&--\\ 
			NovelTagging
			~\citep{zheng2017Joint}
			&--&--&--&59.3&38.1&46.4&46.9&48.9&47.9&\textbf{53.6}&30.3&38.7\\ 
			ReHession
			~\citep{liu2017Heterogeneous}
			&--&--&--&--&--&--&--&--&--&36.7&49.3&42.1\\
			CopyR
			~\citep{zeng2018Extracting}
			&--&--&--&56.9&45.2&50.4&34.7&53.4&42.1&--&--&--\\
			HRL
			~\citep{takanobu2019Hierarchical}
			&--&--&--&71.4&58.6&64.4&\textbf{53.8}&53.8&53.8&--&--&--\\
			PA-LSTM-CRF
			~\citep{dai2019Joint}
			&--&--&--&--&--&--&--&--&--&51.1&39.3&44.4\\
			\midrule
			\textsc{CasRel}
			&57.2&47.6&\textbf{52.0}&\textbf{77.7}&\textbf{68.8}&\textbf{73.0}&50.1&\textbf{58.4}&\textbf{53.9}&49.8&42.7&\textbf{45.9}\\
			\bottomrule 
		\end{tabular} 
	}
	\caption{Relational triple extraction results of different methods under \emph{Partial Match} and \emph{Exact Match} metrics.} 
	\label{tab:more_result} 
\end{table*}

\section{Supplemental Experiments}\label{appendix:sup_exp}
In addition to validating the effectiveness of the proposed \textsc{CasRel} framework in handling the \emph{overlapping triple problem}, we also conduct a set of supplemental experiments to show the generalization capability in more general cases on four widely used datasets, namely, ACE’04, NYT10-HRL, NYT11-HRL and Wiki-KBP. Unlike the datasets we adopt in the main experiment, most test sentences in these datasets belong to the \emph{Normal} class where no triples overlap with each other. Table~\ref{tab:more_result} shows the result of a comprehensive comparison with recent state-of-the-art methods.

\par Notably, there are two different evaluation metrics selectively adopted among previous works: (1)~The widely used one is \emph{\textbf{Partial Match}} as we described in Section~\ref{exp:metric}, i.e., an extracted relational triple \emph{(subject, relation, object)} is regarded as correct only if the relation and the heads of both subject and object are all correct~\citep{li2014Incremental,miwa2016End,katiyar2017Going,zheng2017Joint,zeng2018Extracting,takanobu2019Hierarchical,li2019Entity,fu2019GraphRel}; (2) The stricter but less popular one is \emph{\textbf{Exact Match}} adopted by~\citet{dai2019Joint}, where an extracted relational triple \emph{(subject, relation, object)} is regarded as correct only if the relation and the heads and tails of both subject and object are all correct.

\par Since some works like~\citep{zeng2018Extracting} can't handle multi-word entities and can only be evaluated under the \emph{Partial Match} metric and some works like~\citep{dai2019Joint} are not open-source, it's hard to use a unified metric to compare our model with existing models. To properly compare our model with various baselines, we adopt the \emph{Partial Match} metric for ACE’04, NYT10-HRL and NYT11-HRL, and adopt the \emph{Exact Match} metric for Wiki-KBP.

\paragraph{ACE’04} We follow the same 5-fold cross-validation setting as adopted in previous works~\citep{li2014Incremental,miwa2016End,li2019Entity} and use the code\footnote{https://github.com/tticoin/LSTM-ER} released by~\citep{miwa2016End} to preprocess the raw XML-style data for fair comparison. Eventually, it has 2,171 valid sentences in total and each sentence contains at least one relational triple.

\paragraph{NYT10-HRL \& NYT11-HRL} NYT corpus has two versions: (1) the original version of which both the training set and test set are produced via distant supervision by~\citet{riedel2010Modeling} and (2) a smaller version with fewer relation types, where the training set is produced by distant supervision while the test set is manually annotated by~\citet{hoffmann2011Knowledge}. Here we denote the original one and the smaller one as NYT10 and NYT11, respectively. These two versions have been selectively adopted and preprocessed in many different ways among various previous works, which may be confusing sometimes and lead to incomparable results if not specifying the version. To fairly compare these models, HRL~\citep{takanobu2019Hierarchical} adopted a unified preprocessing for both NYT10 and NYT11, and provided a comprehensive comparison with previous works using the same datasets. Here we denote the preprocessed two versions as NYT10-HRL and NYT11-HRL.

\par For fair comparison, we use the preprocessed datasets released by~\citet{takanobu2019Hierarchical}, where NYT10-HRL contains 70,339 sentences for training and 4,006 sentences for test and NYT11-HRL contains 62,648 sentences for training and 369 sentences for test. We also create a validation set by randomly sampling 0.5\% data from the training set for each dataset as in ~\citep{takanobu2019Hierarchical}.

\paragraph{Wiki-KBP} We use the same version as~\citet{dai2019Joint} adopted, where the training set is from~\citep{liu2017Heterogeneous} and the test set is from~\citep{ren2017Cotype}. It has 79,934 sentences for training and 289 sentences for test. We also create a validation set by randomly sampling 10\% data from the test set as~\citet{dai2019Joint} suggested.

\renewcommand{\arraystretch}{1} 
\begin{table*}[!t] 
	\centering 
	\small
	\begin{tabular}{cccccccc} 
		\toprule
		\multirow{2}{*}{Category}&
		\multicolumn{1}{c}{ACE’04}&
		\multicolumn{2}{c}{NYT10-HRL}&
		\multicolumn{2}{c}{NYT11-HRL}&
		\multicolumn{2}{c}{Wiki-KBP}
		\\
		\cmidrule(lr){2-2}\cmidrule(lr){3-4}\cmidrule(lr){5-6}\cmidrule(lr){7-8}
		&ALL&Train&Test&Train&Test&Train&Test\\
		\midrule 
		\textit{Normal}&1604&59396&2963&53395&368&57020&265\\
		\textit{EPO}&8&5376&715&2100&0&3217&4\\
		\textit{SEO}&561&8772&742&7365&1&21238&20\\
		\midrule 
		ALL&2171&70339&4006&62648&369&79934&289\\
		\bottomrule 
	\end{tabular} 
	\caption{Statistics of datasets. Note that a sentence can belong to both \emph{EPO} class and \emph{SEO} class.} 
	\label{tab:statistics2} 
\end{table*}

\paragraph{Dataset Study} As stated beforehand, these datasets are not suitable for testing the overlapping problem. To further explain this argument, we analyze the datasets in detail and the statistics are shown in Table~\ref{tab:statistics2}. We find that the test data in these datasets suffer little from the so-called \emph{overlapping triple problem} since the sentences contain few overlapping triples. Even worse, we also find that the annotated triples are not always the ground truth, which may affect the evaluation of our model. Take a real instance from the test set of NYT11-HRL for example, for the sentence \emph{\small ``Mr. Gates , who arrived in Paris on Tuesday , was the first American defense secretary to visit France in nearly a decade ."}, the annotated triple set only contains one relational triple:\\ \emph{\small \{(Paris, /location/administrative\_division/country, France)\}}. However, the output of our model has three triples: \emph{\small \{(Paris, /location/administrative\_division/country, France), (France, /location/country/administrative\_divisions, Paris), (France, /location/location/contains, Paris)\}}. The last two triples should have been annotated in the sentence but were omitted, which will significantly affect the values of both precision and recall when quantifying the performance of our model.

\par This observation demonstrates that our \textsc{CasRel} model could extract more relational triples than the manually annotated ones in some cases due to the imperfect annotation. For this reason, the performance on such dataset like NYT11-HRL can only partially reflect the potential of the proposed model and probably underestimate the real value. Nonetheless, the \textsc{CasRel} model still achieves a competitive performance, showing the effectiveness of the proposed novel cascade binary tagging framework in relational triple extraction.

\par We also note that there is a significant gap (from 53.9 to 89.6 in terms of F1-score) between the performance on NYT11-HRL that preprocessed by~\citet{takanobu2019Hierarchical} and on NYT11-CopyR\footnote{We denote the version preprocessed by~\citet{zeng2018Extracting} as NYT11-CopyR for disambiguation, which is also referred to as ``NYT"  in the main experiment.} that preprocessed by~\citet{zeng2018Extracting}. Though both of the above two versions are adapted from the original NYT11 dataset~\cite{hoffmann2011Knowledge}, there are two key differences in the NYT11-CopyR version as~\citet{dai2019Joint} pointed out. First, instead of using the manually annotated test set, ~\citet{zeng2018Extracting} randomly select 5000 sentences from the training data as the test data. The reason is that the manually annotated data contains few overlapping triples and thus not suitable for testing the overlapping triple problem; Second,~\citet{zeng2018Extracting} only annotated the \textbf{last word} of an entity in both training and test sets because their model can not handle multi-word entities. Hence, any entity in their dataset is taken as a single-word entity, leading to that the \emph{Partial Match} and \emph{Exact Match} evaluation metrics make no difference. Moreover, such setting makes it much easier for our \textsc{CasRel} model to detect the span of an entity since the start and end positions are actually the same. We attribute the significant gap to these different settings between the above two versions. Noticeably, multi-word entities are common in real-world scenarios, so evaluating on a more proper dataset like NYT10-HRL (which also contains overlapping triples in test set) may better reveal the model's real value in relational triple extraction than on the ad-hoc one.